\documentclass[conference]{IEEEtran}
\IEEEoverridecommandlockouts
\usepackage{caption}
\usepackage{subcaption}
\usepackage{cite}
\usepackage{amsmath,amssymb,amsfonts}
\usepackage[ruled,vlined]{algorithm2e}
\usepackage{graphicx}
\usepackage{textcomp}
\usepackage{xcolor}
\usepackage{multirow}
\usepackage[normalem]{ulem}
\usepackage{float}
\useunder{\uline}{\ul}{}

\begin{document}

\title{Scalable Multiagent Reinforcement Learning with Collective Influence Estimation}

\author{Zhenglong Luo, Zhiyong Chen, Aoxiang Liu, and Ke Pan
\thanks{Z. Luo and Z Chen are with the School of Engineering, The University of Newcastle, Callaghan, NSW 2308, Australia. 
A. Liu and K. Pan are with the School of Automation, Central South University, Changsha 410083, China.
Z. Chen is the corresponding author. E-mail: zhiyong.chen@newcastle.edu.au. }}

\maketitle

\begin{abstract}
Multiagent reinforcement learning (MARL) has attracted considerable attention due to its potential in addressing complex cooperative tasks. However, existing MARL approaches often rely on frequent exchanges of action or state information among agents to achieve effective coordination, which is difficult to satisfy in practical robotic systems. A common solution is to introduce estimator networks to model the behaviors of other agents and predict their actions; nevertheless, such designs cause the size and computational cost of the estimator networks to grow rapidly with the number of agents, thereby limiting scalability in large-scale systems.

To address these challenges, this paper proposes a multiagent learning framework augmented with a Collective Influence Estimation Network (CIEN). By explicitly modeling the collective influence of other agents on the task object, each agent can infer critical interaction information solely from its local observations and the task object's states, enabling efficient collaboration without explicit action information exchange. The proposed framework effectively avoids network expansion as the team size increases; moreover, new agents can be incorporated without modifying the network structures of existing agents, demonstrating strong scalability. Experimental results on multiagent cooperative tasks based on the Soft Actor–Critic (SAC) algorithm show that the proposed method achieves stable and efficient coordination under communication-limited environments. Furthermore, policies trained with collective influence modeling are deployed on a real robotic platform, where experimental results indicate significantly improved robustness and deployment feasibility, along with reduced dependence on communication infrastructure.
\end{abstract}


\section{Introduction}

Reinforcement Learning (RL) was traditionally devised to tackle single-agent decision-making challenges, with the primary goal of enhancing the behavior of an individual agent through interaction with its environment. Early approaches, exemplified by Q-learning \cite{watkins1992q}, depend on tabular value function representations and are effective primarily in small-scale state–action spaces, with their scalability diminishing rapidly as problem complexity grows. To address this limitation, Deep Q-Networks (DQN) \cite{lecun2015deep} employed neural networks as function approximators, allowing reinforcement learning to manage high-dimensional state spaces. Nonetheless, value-based approaches continue to be constrained in continuous action control tasks and are frequently challenging to implement directly within robotic systems.

Building upon these advancements, research has gradually transitioned toward policy-driven and hybrid methodologies. The Actor–Critic (AC) framework \cite{KondaTsitsiklis2000} integrates policy optimization with value estimation, effectively maintaining an advantageous balance between learning efficiency and training stability. Representative algorithms such as Twin Delayed Deep Deterministic Policy Gradient (TD3) \cite{fujimoto2018td3} and Soft Actor–Critic (SAC) \cite{haarnoja2018sac} further improve stability and exploration in continuous-control tasks through the integration of twin critic architectures, delayed policy updates, and entropy regularization. These approaches now form the fundamental basis of contemporary continuous-control reinforcement learning.

Nevertheless, as reinforcement learning techniques are more frequently employed in complex real-world systems, the shortcomings of the single-agent assumption have become apparent. In practical applications, systems frequently consist of multiple agents that are required to interact, collaborate, or compete within dynamic environments. Single-agent reinforcement learning frameworks lack the capacity to explicitly model inter-agent interactions and are ineffective in addressing the non-stationarity and decision coupling resulting from multiagent dynamics. This practical necessity has directly driven the advancement of multiagent Reinforcement Learning (MARL), which seeks to develop intelligent systems capable of effective coordination and adaptive decision-making within multiagent environments.

In transitioning reinforcement learning from single-agent to multiagent environments, a primary challenge involves modeling and leveraging interactions among agents, thereby rendering information-sharing mechanisms a critical component in the design of MARL algorithms. In multiagent reinforcement learning scenarios, the method by which agents exchange information directly influences policy assessment, collaborative effectiveness, and the overall scalability of the system. A substantial portion of existing MARL approaches depend on explicit inter-agent communication to facilitate cooperation and evaluate strategies. Representative works include Nash Q-learning \cite{hu2003nash}, which integrates Nash equilibria into Q-value updates to model strategic interdependencies among agents; the subsequent Diverse Q-Vectors approach \cite{luo2025multi}, which learns multiple game-theoretic Q-vectors (e.g., Max, Nash, and Maximin) within a deep network framework, offering a unified value-function perspective for multiagent decision-making; as well as centralized-training–decentralized-execution paradigms such as multiagent Deep Deterministic Policy Gradient (MADDPG) \cite{lowe2017multi} and Counterfactual multiagent Policy Gradient (COMA) \cite{foerster2018counterfactual}. Additionally, communication and structure-oriented deep multiagent reinforcement learning frameworks have been extensively investigated, including Differentiable Inter-Agent Learning (DIAL) \cite{foerster2016dial}, Communication Networks (CommNet) \cite{sukhbaatar2016commnet}, and graph neural network (GNN)-based approaches \cite{jiang2020graph}. While communication-oriented multiagent reinforcement learning techniques can markedly improve training stability and collaborative efficacy through the exchange of global or local information, they also lead to heightened computational complexity, augmented communication bandwidth demands, and intensified system interdependence. Consequently, these methodologies encounter significant scalability constraints when implemented in extensive systems and practical engineering applications.

To resolve these issues, there has been a growing focus on communication-limited or communication-free MARL frameworks, which seek to minimize communication overhead while preserving system stability. Direct methodologies, including Independent Q-Learning (IQL) and Independent Policy Gradient (IPG) \cite{tan1993iql}, regard other agents as components of the environment; nonetheless, these techniques frequently encounter convergence challenges or produce suboptimal outcomes in practice. Opponent modeling and temporal modeling techniques have been implemented to address these deficiencies. For example, Learning with Opponent-Learning Awareness (LOLA) \cite{foerster2018lola} explicitly models the policy gradients of opponents to facilitate proactive policy adaptation, whereas Deep Recurrent Q-Networks (DRQN) \cite{hausknecht2015deep} utilize recurrent neural networks to capture temporal dependencies, thus enhancing performance in conditions of partial observability.

Collectively, these methods demonstrate the feasibility of modeling other agents from diverse perspectives, offering effective means to characterize multiagent interactions without explicit communication and further advancing estimation-network–based multiagent reinforcement learning. The primary concept of these methodologies is to explicitly or implicitly deduce the behavioral information of other agents—without depending on direct communication and exclusively relying on each agent's individual observations—to tackle non-stationarity and partial observability in communication-restricted multiagent reinforcement learning settings. Among these, the Action Estimation Network (AEN)\cite{luo2026aen} functions as a representative approach, explicitly predicting the actions of other agents based solely on locally observable information and integrating the estimated actions into value function approximation or policy learning. By doing so, AEN substantially improves the stability and resilience of cooperative multiagent learning in communication-limited environments.

Similarly, Regularized Opponent Model Multi-Agent Reinforcement Learning (ROMMEO) \cite{yang2020rommeo} approximates static opponent policies via regularized optimization, whereas Bayesian Opponent Learning Optimization (BOLO) \cite{albrecht2018bolo} utilizes Bayesian inference to continuously maintain and update a posterior distribution over opponent policies in real time. Opponent Modeling with Temporal Convolutional Networks (OM-TCN) \cite{du2021omtcn} captures the temporal evolution of opponent behaviors using temporal convolutional structures. In addition, DPIQN/DRPIQN \cite{hong2018dpiqn,hong2019drpiqn}, Probabilistic Recursive Reasoning (PR2) \cite{wen2019pr2}, Multi-Agent Offline Policy Inference (MAOP) \cite{zhang2021maop}, and Imitated Opponent Q-learning (I2Q) \cite{wang2020i2q} model multiagent interactions from the perspectives of policy embeddings, recursive reasoning, or future state estimation. Although these approaches partially mitigate non-stationarity and partial observability in communication-limited MARL, they typically rely on explicit opponent modeling, complex temporal or recursive architectures, or extensive historical interaction data. As the number of agents and task complexity increase, such dependencies often lead to rapidly growing model complexity, unstable training dynamics, and limited real-time decision-making capability, thereby restricting their applicability in communication or perception limited engineering environments. Consequently, effectively characterizing inter-agent influence without incurring additional communication overhead or excessive model complexity remains a key open challenge in communication-limited multiagent reinforcement learning.

To address cooperative tasks within these constraints, drawing inspiration from the aggregation modeling principle in mean-field reinforcement learning \cite{yang2018meanfield} and the action estimation capabilities exhibited by AEN-TD3 and OM-TCN, this paper introduces a straightforward and scalable estimation module named the Collective Influence Estimation Network (CIEN). Unlike AEN, CIEN does not necessitate the direct supervision of additional agents. It depends solely on the individual state of each agent and the condition of the shared object. By documenting the evolutionary trajectories of the shared object during exploration, CIEN acquires the ability to encapsulate the cumulative impact of teammates on the object's ensuing dynamics. The estimated collective influence is integrated into the actor network as an environmental input, facilitating more precise action selection without the need for explicit inter-agent observation or communication.

A primary advantage of this design is that the input–output dimensionality of the collective influence estimation network remains constant regardless of the number of agents, thereby effectively preventing the dimensionality explosion commonly encountered in large-scale multiagent systems. This invariance confers substantial scalability to the proposed approach: upon the addition of new agents, neither the environment nor the network architecture necessitates alteration, and each agent can readily duplicate the existing network and commence training following initialization. Furthermore, throughout both training and execution phases, the estimation network solely relies on the states of individual agents and the historical states of the shared object, thereby minimizing dependence on communication.

To assess the efficacy of the proposed framework, comprehensive experiments are performed in the MuJoCo simulator \cite{todorov2012mujoco} involving a three-arm collaborative manipulation task modeled under realistic physical conditions. The experimental configuration includes frictional effects, gravitational forces, and vibrational disturbances generated by joint interactions, rendering the task closely analogous to real-world multiagent robotic collaboration scenarios. To enhance the evaluation of deployability, the trained policies are transferred to actual robotic platforms for validation.

Experimental results demonstrate that CIEN-SAC displays significant robustness against typical non-idealities and uncertainties in real robotic systems, emphasizing its considerable potential for applying MARL in environments with communication limitations and dynamic complexity.

\section{Methods}

In this section, we systematically introduce the proposed method by focusing on the core issues of Scalable Multiagent Reinforcement Learning. First, we present the modeling concept of Collective Influence and, based on this, introduce the Collective Influence Estimation Network (CIEN). This network aims to estimate the collective influence of other agents on task-relevant state evolution during multiagent collaboration, enabling stable and efficient cooperative learning without explicit communication.

At the implementation level, CIEN is designed as a modular, algorithm-agnostic estimation component that can be flexibly integrated into various reinforcement learning frameworks, such as Actor–Critic (AC), Deep Q-Network (DQN), and Soft Actor–Critic (SAC). When combined with SAC, the resulting integrated algorithm is termed CIEN-SAC. Tailored for scalable multiagent reinforcement learning scenarios, this algorithm is particularly effective in environments where agents cannot exchange action information or face communication constraints. By estimating collective influence, it significantly enhances the stability, scalability, and collaborative performance of policy learning.
 
\subsection{Collective Influence Modeling}

In cooperative multiagent reinforcement learning tasks, the development of shared task objectives or environment-dependent critical states is collectively influenced by the actions of multiple agents. From a Markov Decision Process (MDP) perspective, when the states or actions of other agents are unobservable or inaccessible due to communication limitations, their decisions are no longer represented as explicit variables but are implicitly incorporated into the environment's state transition dynamics. From the perspective of a single agent, the actions of teammates thus appear as unpredictable and non-stationary elements within the environment, representing a primary source of instability in multiagent learning without communication.

Based on this observation, we propose the concept of Collective Influence, which delineates the cumulative effect of the combined behaviors of other agents on the progression of task-relevant states. Instead of explicitly modeling the actions or policies of individual agents, Collective Influence approximates the aggregate effect of all other agents from the viewpoint of a specific agent as a low-dimensional, quantifiable influence factor. 

\begin{figure}
     \centering
     \includegraphics[width=0.45\textwidth]{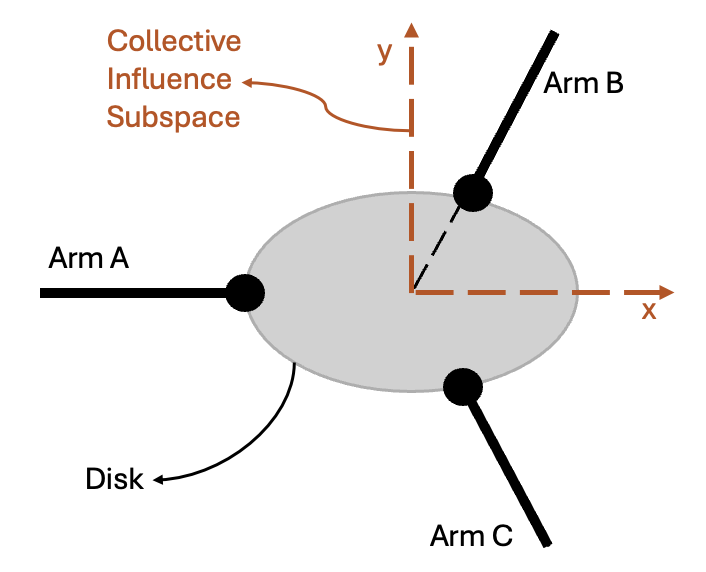}
     \caption{Three-arm collaborative robotic system}
     \label{CI}
\end{figure}

For example, in the three-arm collaborative robotic system shown in Figure~\ref{CI}, each robotic arm controls three joints. From the perspective of Arm A, traditional adversarial modeling–based estimation networks typically require explicit estimation of the motion information of the other two arms, resulting in an estimation dimension of $3 \times 2 = 6$. In contrast, under the collective influence modeling framework, the combined effect of the other two arms on the collaborative task can be approximated as a synthetic influence that lies within a low-dimensional Collective Influence Subspace. Specifically, the positional distribution of this synthetic influence relative to the shared object and Arm A exhibits a symmetric structure (as illustrated by the dashed region in the figure). Under this approximation, the collective influence primarily manifests as the displacement variation of the shared object within the $x$–$y$ plane induced by the other two arms. Consequently, this influence can be characterized by only two variables within the Collective Influence Subspace, thereby significantly reducing the estimation dimension to $2$.

This formulation allows a learning agent to consider multiagent interactions without the need for explicit inter-agent communication or extensive opponent modeling. Furthermore, by consolidating multiagent effects into a concise representation with a dimensionality that is independent of the number of agents, Collective Influence can be derived exclusively from task-level observable data—such as shared object dynamics—rendering it especially suitable for scalable and resilient cooperation in communication-limited multiagent settings.

\subsection{Estimation Network}
\label{subsec:CIEN}

A reinforcement learning algorithm updates the action-value function $Q(s, a)$ through the transition tuple $(s, a, r, s_{\text{next}}, a_{\text{next}})$, where $s$ denotes the current state, $a$ is the action taken under the current policy, $r$ is the immediate reward, $s_{\text{next}}$ is the next state, and $a_{\text{next}}$ is the subsequent action selected by the same policy.

In the multiagent scenario studied here, the reward $r$ is defined for a cooperative target. For example, in the three-arm lifting task, the reward is determined by the height $r_{\rm height}$ of the object (e.g., a disk) and the posture-related term $r_{\rm angle}$ (defined as the negative tilt angle), such that
$r = r_{\rm height} + r_{\rm angle}$. The robots cooperate to maximize the accumulated reward.

Each agent learns its policy independently in a decentralized manner. In most multiagent algorithms, for agent $i$, its state is denoted by $s^i$, and the states of the remaining agents are denoted by $s^{-i}$. Thus, the overall state can be expressed in compound form as $s = (s^i, s^{-i})$. Similarly, the action is decomposed as $a = (a^i, a^{-i})$, where $a^i$ is the action of agent $i$, and $a^{-i}$ represents the actions of all other agents. Since the primary source of uncertainty in a multiagent environment arises from the actions taken by other agents, this definition best preserves the true state transition process, making learning more accurate when both states and actions are observable.

However, in most practical scenarios, although an agent can access its own full state as well as the state of the shared object, it is generally unable to observe the states of other agents, denoted as $s^{-i}$—a limitation that becomes particularly pronounced when the number of agents is greater than or equal to three. Moreover, when the environment does not support direct inter-agent communication, the joint action $a = (a^i, a^{-i})$ becomes incomplete, containing only the agent’s own action $a^i$ while the actions of other agents $a^{-i}$ are missing.

As a result, from the perspective of agent $i$, the action information provided to the critic network is incomplete, which increases the estimation error of the Q-value. While employing an Action Estimation Network (AEN) to infer other agents’ actions $a^{-i}$ is feasible in two-agent environments, as the number of agents increases, the output dimensionality of action estimation networks grows rapidly, which significantly limits the scalability of such methods in  multiagent systems.

Therefore, instead of explicitly estimating other agents' actions $a^{-i}$, we estimate the joint effect introduced earlier—namely, the collective influence $c^{-i}$—and adopt a more advanced Collective Influence Estimation Network (CIEN) to infer this quantity. The concept is illustrated as follows:
\begin{align}
c^{-i} = e_{\psi}(s^{o})  
\end{align}
Here, $s^{o}$ denotes the state of task object, for instance, the disk's height and angle in the lift task, and $e_{\psi}$ denotes a neural network parameterized by the weight vector $\psi$. Similar to the critic network in SAC, CIEN also maintains a target network, denoted as $e_{\psi^{\prime}}(s^{o})$, which is parameterized by a separate weight vector $\psi^{\prime}$.

CIEN-SAC adopts the standard Double Q-learning algorithm, which employs two independently trained critic networks to estimate Q-values, denoted as $Q_{\theta_1}(s^i, s^o, a^i, c^{-i})$ and $Q_{\theta_2}(s^i, s^o, a^i, c^{-i})$. These networks are parameterized by the weight vectors $\theta_1$ and $\theta_2$, respectively, and have corresponding target networks with parameters $\theta_1^{\prime}$ and $\theta_2^{\prime}$. This approach is designed to alleviate overestimation bias in value estimation.

With the incorporation of the CIEN module, the fundamental distinction between these critic networks and their conventional counterparts lies in the fact that Q-value estimation relies on the state of task object's state $s^o$ together with the collective influence $c^{-i}$, rather than on the other agents $s^{-1}$ associated with the states of the actual joint action component $a^{-i}$. 

\begin{figure}
    \centering
    \includegraphics[width=0.45\textwidth]{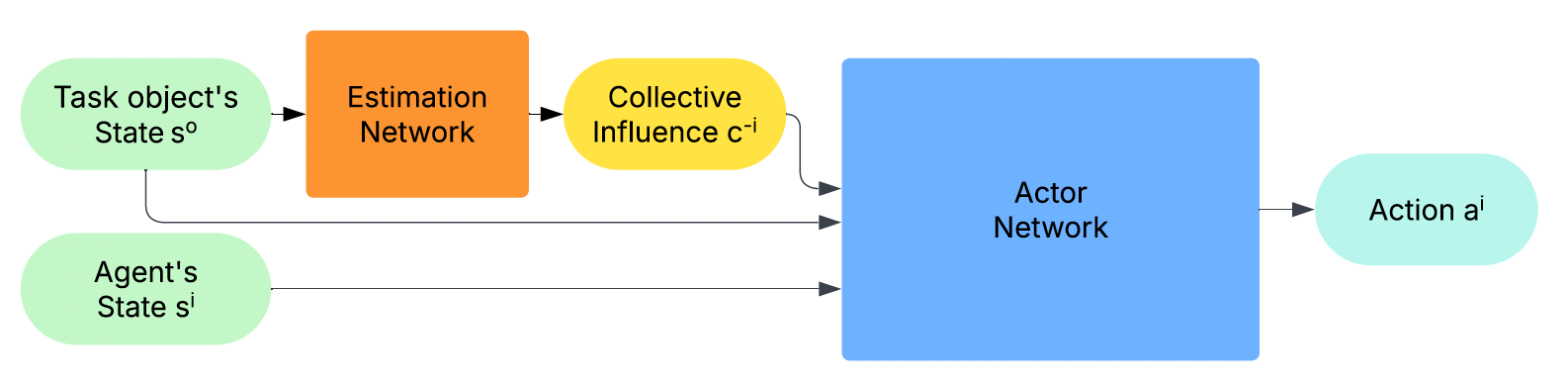}
    \caption{Flowchart of the Actor network with the CIEN.}
    \label{flowchart}
\end{figure}

In a conventional centralized setting, the actor network is represented as $a = \pi_\phi(s^i, s^o, c^{-i})$, which generates the complete action vector for all agents. In contrast, under the decentralized setting considered here, the state of the manipulated object is first processed through the CIEN module to estimate the collective influence $c^{-i}$ of the other agents. Then, the actor network of agent $i$ generates its own action $a^i$ based on its individual state and the collective object state $s^o$, jointly denoted as $s^i$, together with the estimated collective influence $c^{-i}$. The overall procedure is illustrated in Figure~\ref{flowchart}. The actor network is parameterized by the weight vector $\phi$, and due to the particular characteristics of SAC, no target actor network is maintained.

\subsection{Action Selection}

In the CIEN-SAC framework, the policy $\pi_\phi(a^i \mid s^i, s^o, c^{-i})$ is modeled as a stochastic distribution, typically parameterized as a Gaussian with mean and variance predicted by the actor network. This formulation allows SAC to naturally incorporate exploration through sampling, rather than relying on external noise processes as in DDPG or TD3.  

Given a state $s$ and collective influence $c^{o}$, the actor network outputs the parameters of the Gaussian distribution, $\mu_\phi(s^i, s^o, c^{-i})$ and $\sigma_\phi(s^i, s^o, c^{-i})$. An action is then sampled using the reparameterization trick:  
\begin{align}
a^{i} & = \pi_{\phi}(s^i, s^o, c^{-i}) \nonumber \\
&= \tanh\big(\mu_\phi(s^i, s^o, c^{-i}) + \sigma_\phi(s^i, s^o, c^{-i}) \odot \epsilon\big), 
\end{align}
where the $\tanh$ function ensures that the resulting action lies within the valid range $[a_{\min}, a_{\max}]$, and 
$\epsilon \sim \mathcal{N}(0,I)$  denotes Gaussian noise introduced via the reparameterization trick

This stochastic action selection mechanism not only promotes efficient exploration but also enables the maximum entropy objective of SAC, which encourages the policy to maintain sufficient randomness. The trade-off between exploration and exploitation is governed by the temperature parameter $\alpha$, which weights the entropy term in the objective function.

\subsection{Update of Networks}

The entropy regularized temporal difference target used for updating the critic networks is defined as follows: 
\begin{align}
y =& r + \gamma \min_{j=1,2} Q_{\theta'_j}(s^{i}_{\text{next}}, s^{o}_{\text{next}}, a^{i}_{\text{next}}, c^{-i}_{\text{next}}) \nonumber \\
&+ \alpha \log \pi_{\phi}(a^{i}_{\text{next}} \mid s^{i}_{\text{next}}, s^{o}_{\text{next}}, c^{-i}_{\text{next}})),  \nonumber \\
a^{i}_{\text{next}} =& \pi_{\phi}(s^{i}_{\text{next}}, s^{o}_{\text{next}}, 
c^{-i}_{\text{next}}), \quad
c^{-i}_{\text{next}} = e_{\psi^{{\prime}}}(s^{o}_{\text{next}}),
\label{eq:sac_q_target}
\end{align}
where $\gamma$ is the discount factor, $s^{i}_{\text{next}}$ denotes the state of agent $i$, and $s^{o}_{\text{next}}$ characterizes the state of task object {o}. The term $\alpha \log \pi_{\phi}(a^{i}_{\text{next}} \mid s^{i}_{\text{next}}, s^{o}_{\text{next}}, c^{-i}_{\text{next}})$ acts as entropy regularization, balancing exploration and exploitation. A larger $\alpha$ emphasizes exploration by encouraging diverse action selection, while a smaller $\alpha$ prioritizes exploiting high-value actions, thus preventing premature convergence.

In \eqref{eq:sac_q_target}, the actor network $\pi_{\phi}$ and the target CIEN $e_{\psi^{\prime}}$ are used to generate the next predicted actions for agent $i$ and collective influence $c^{-i}_{\text{next}}$ by other agents, respectively. The target Q-value is then computed using the target critic networks $Q_{\theta'_j}$, for $j=1,2$. The loss for each critic network $Q_{\theta_j}$ ($j=1,2$) is defined as: 
\begin{align}
L(\theta_j) = \mathbb{E}_{(s^i, s^o, a^{i}, c^{-i}, r, s^{i}_{\text{next}},s^{o}_{\text{next}})\sim\mathcal{B}}
\nonumber\\
\left[(Q_{\theta_j}(s^i, s^o, a^{i}, c^{-i}) - y)^2\right], \label{loss}
\end{align}
where $\mathcal{B}$ denotes the replay buffer from which experience tuples are sampled. The critic network parameters $\theta_1$ and $\theta_2$ are updated by minimizing the corresponding loss functions.

To update the actor network, CIEN-SAC adopts the same delayed policy update strategy as SAC, where the policy network $\pi_{\phi}$ is updated less frequently than the critic networks. Specifically, the actor and its corresponding target network are updated once every $d$ critic updates, with $d$ typically set to $2$. During each update, the objective is to maximize the Q-value estimated by the first critic network $Q_{\theta_1}$. Accordingly, the actor parameters $\phi$ are updated by applying the following gradient:
\begin{align}
\nabla_{\phi}J(\phi) = \mathbb{E}_{(s^i, s^o, a^{i}, c^{-i}, r, s^{i}_{\text{next}}, s^{o}_{\text{next}})\sim\mathcal{B}} \nonumber\\
\left[\nabla_{\phi}(\alpha \log \pi_{\phi}(a^{i} \mid s^i, s^o, c^{-i}) - Q_{\theta_1}(s^i, s^o ,a^{i}, c^{-i})))\right].
\label{actorpdate}
\end{align}

Since the CIEN essentially serves to estimate the actions of other agents, it can be regarded as a form of policy approximation, and its neural network architecture is therefore similar to the actor network in Actor–Critic methods. However, given that the actor networks of other agents incorporate policy entropy during updates when trained with the same algorithm, the CIEN also retains policy entropy in its updates in order to ensure estimation accuracy. Specifically, the parameters of CIEN, denoted by $\psi$, are updated according to the following gradient:
\begin{align}
\nabla_{\psi}J(\psi) = \mathbb{E}_{(s^i, s^o, a^{i}, c^{-i}, r, s^{i}_{\text{next}}, s^{o}_{\text{next}})\sim\mathcal{B}} \nonumber\\
\left[\nabla_{\psi}(\alpha \log \pi_{\psi}(a^{i} \mid s^i, s^o, c^{-i}) - Q_{\theta_1}(s^i, s^o, a^{i}, c^{-i})))\right].
\label{cienupdate}
\end{align}

After updating the network parameters $\theta_1$, $\theta_2$, $\phi$ and $\psi$, the corresponding target networks are updated using a soft update mechanism:
\begin{align}
      &\theta'_j \leftarrow \tau \theta_j + (1 - \tau) \theta'_j,\; j=1,2 \nonumber\\
      &\phi' \leftarrow \tau \phi + (1 - \tau) \phi' \nonumber\\
      &\psi' \leftarrow \tau \psi + (1 - \tau) \psi' 
\label{targetupdate}
\end{align}
where $\tau \in (0, 1)$ is the soft update rate that controls the speed of target network tracking.

\begin{algorithm}[t]
\caption{CIEN-SAC Algorithm for Agent $i$}
\label{alg:sac}

\KwSty{Parameter Setting}: Total episodes $M$, episode length $T$, minibatch size $N$, entropy coefficient $\alpha$, soft update rate $\tau$\;

\KwSty{Initialize networks}: Initialize the critic networks  $\theta_1$ and $\theta_2$, the actor network $\phi$, and the CIEN $\psi$\;

\KwSty{Initialize target networks}: $\theta^{\prime}_1 \leftarrow \theta_1$, 
$\theta^{\prime}_2 \leftarrow \theta_2$, $\psi^{\prime} \leftarrow \psi$\;

\KwSty{Initialize replay buffer} $\mathcal{B}$\;

\For{episode $= 1$ to $M$}{
  Observe initial state $s_1^i, s_1^{o}$ \;
  \For{$t = 1$ to $T$}{
    Estimate collective influence: $c^{-i}_t =  e_{\psi}(s^{o}_t)$\;
    Select action: $a^i_t = \pi_{\phi}(s^i_t,s^o_t, c^{-i}_t)$\;
    Execute action $a^i_t$, observe reward $r_t$ and next state 
    $s_{t+1}^i, s_{t+1}^{o}$\;
    Store transition $(s^i_t,s^o_t, a^i_t, c^{-i}_t, r_t, s_{t+1}^i, s_{t+1}^{o})$ in $\mathcal{B}$\;
    Sample a minibatch of $N$ transitions $(s^i,s^o, a^i, c^{-i}, r, s^i_{\text{next}},s^o_{\text{next}})$ from $\mathcal{B}$\;
    Calculate the TD target with entropy term $y$ for each sample according to
    \eqref{eq:sac_q_target}\;

    Compute the empirical mean of the loss \eqref{loss}:
    $L(\theta_j) = \frac{1}{N} \sum (Q_{\theta_j}(s^i,s^o, a^i, c^{-i})-y )^2$, $j=1,2$\;

    Update the critic networks $\theta_1$ and $\theta_2$ by minimizing their corresponding losses\; 

    Update the actor network $\phi$ by applying the empirical gradient of \eqref{actorpdate}:
      $\nabla_{\phi} J(\phi) = \frac{1}{N} \sum\nabla_{\phi}Q_{\theta_1}(s^i,s^o,\pi_{\phi}(s^i,s^o, c ^{-i}), c^{-i}))$\;

      Update the CIEN $\psi$ by applying the empirical gradient of \eqref{cienupdate}:
      $\nabla_{\psi} J(\psi) = \frac{1}{N} \sum \nabla_{\psi}Q_{\theta_1}(s^i,s^o, a^i, e_{\psi}(s^{o}))$\;

      Update the target networks according to \eqref{targetupdate}.
    
  }
}
\end{algorithm}

With the network architecture and update rules introduced above, the overall workflow of the proposed CIEN-SAC algorithm is summarized in Algorithm~\ref{alg:sac}. The method integrates CIEN into the standard SAC framework, enabling an aggregated modeling of the collective influence of other agents on the cooperative task without explicitly recovering their individual policies or actions. Owing to this design, CIEN-SAC can provide the critic network with more informative inputs even in the complete absence of inter-agent communication, thereby effectively mitigating the non-stationarity inherent in multiagent environments.

Moreover, since the representation of collective influence is invariant to the number of agents, the proposed approach avoids the rapid growth of model complexity and hardware burden as the agent population scales. As a result, in scalable multiagent cooperation scenarios, CIEN-SAC can significantly improve the accuracy of Q-value estimation using a relatively lightweight network structure, while substantially reducing communication dependency and computational resource consumption, and achieving performance comparable to that of centralized training methods.
 
\section{Experiments}
\label{sec.Experiments}

To evaluate the effectiveness of CIEN-SAC in cooperative tasks, we compared its performance against original centralized SAC and enhanced distributed SAC (the latter incorporating state information of the task object—the disk $s^o$). Centralized SAC serves as the baseline approach, while distributed SAC provides a comparative study integrating object state. Experiments were designed and modeled within the Robosuite environment\cite{zhu2020robosuite}, with the task defined as “three robotic arms lifting of a disk.” Following training, both the centralized SAC strategy and the CIEN-SAC strategy were deployed on a physical robot platform to evaluate their real-world deployability. The three robotic arms used in the task were all UR5e models, consistent with the equipment employed in the Robosuite simulation environment.

\subsection{Simulation Experimental Environment}

The three-arm environment is constructed by extending an existing dual-arm lifting task in the Robosuite framework. 
The task objective is to lift a disk-shaped object from an initial height of $0.896\,\mathrm{m}$ to a target height of $1.36\,\mathrm{m}$ above the ground. Specifically, a third same robotic arm is introduced into the original symmetric dual-arm configuration, with the angular separation between each pair of arms adjusted to $120^\circ$. In addition, since the original task object used in the dual-arm setting cannot ensure balanced force distribution under the three-arm configuration, the task object is replaced with a centrally located disk equipped with a ring-shaped handle. 

Apart from adding the third arm and modifying the task object, the grippers were kept at their default settings. The parameter {\it `has renderer'} was set to {\it `True'} to enable real-time visualization of robotic arm movements. The parameter control frequency, which determines the frequency of the input signal, was set to control frequency=4. 

The parameter horizon, which specifies the end of the current cycle after a given number of actions are executed and the environment is reinitialized, was set to {\it `horizon=200'}. These configurations allowed the robotic arms to lift the disk to the target position along the desired trajectory, thereby ensuring effective training completion.
 
In the Robosuite \emph{three-arm disk-lifting} environment, a multi-agent scenario is constructed using three robotic manipulators. For clarity, the three arms are indexed as $i = 1, 2, 3$. The primary objective of the task is to coordinate the joint motions of all three arms such that they collaboratively lift a centrally located disk while maintaining its stability throughout the lifting process. The state of each robotic arm is denoted as $s_t^i \in \mathbb{R}^{6}$, $i = 1, 2, 3$, which consists of the six joint angles of the corresponding manipulator.
In addition, the state of the disk is included as part of the environment observation. The reward function $r_t^i$ is defined in Section~\ref{subsec:CIEN}.
Joint velocities are used as control inputs, with the control frequency set to {\it 'control frequency = 4'}. Specifically, control commands are applied to three joints of each robotic arm at a frequency of $4\,\mathrm{Hz}$, resulting in a continuous action space for each individual arm.

The termination condition for each episode is specified as follows: the environment is reset after a fixed number of action executions, with the episode horizon set to {\it 'horizon = 200'}.
This setting allows the three robotic arms to lift the disk to the target height while following an approximately optimal trajectory, thereby ensuring effective task completion during training.

To facilitate safe transfer of the learned policies to real robotic systems, additional early termination conditions are introduced based on the detection of unsafe mechanical configurations. These early termination conditions enable the episode to be terminated before reaching the full horizon whenever potentially hazardous behaviors are identified.

\begin{enumerate}
    \item \textbf{Gripper to Disk Center Distance:}  
    To prevent excessive squeezing or pulling, the relative distance between each gripper and the center of the disk is monitored throughout the episode. Here, $d_t^{i}$ denotes the distance between the $i$'s gripper and the disk center at time step $t$, while $d_0$ represents the initial distance when the disk is symmetrically grasped. A safe deviation threshold $\delta$ is defined, and early termination is triggered if $|d_t - d_0| > \delta$.
     
    \item \textbf{Disk Center Deviation from Initial Position:}  
    As an auxiliary constraint to the distance monitoring, a safety deviation threshold is introduced to prevent agents from learning harmful force application strategies during training. This criterion is designed to enhance the safety and reliability of the learned policies when deployed on real robotic platforms.

    \item \textbf{Disk Tilt Angle:}  
    To ensure a stable lifting process, an additional constraint is imposed on the disk orientation. Specifically, if the tilt angle of the disk exceeds a predefined threshold $\pi/4$, the current episode is considered a failure and terminated. 
    
\end{enumerate}

If any of the above safety conditions is violated, the episode is regarded as unsafe and terminated prematurely.
Together with the maximum episode length of 200 time steps, these criteria jointly define the termination condition of the environment. Once a termination condition is triggered, the current episode ends, the environment is reset, and the next training episode begins.

\subsection{Simulation Results and Evaluation}

To evaluate the feasibility of the proposed method under communication-limited conditions, three sets of experiments were conducted in the three-arm lifting environment.  First, a standard SAC algorithm was applied under a full communication setting as the baseline. Second, under the same environment but with inter-agent communication disabled, the proposed CIEN-SAC algorithm was evaluated. Finally, to verify that the performance improvement of CIEN-SAC does not solely originate from introducing task object state information as an additional input, a baseline decentralized SAC variant augmented with disk's state observations was implemented for ablation analysis.

\subsubsection{Centrelized Setting: SAC}

In this experiment, the three robotic arms operate under a fully centralized information-sharing setting, where states, actions, rewards, and replay buffers are entirely shared. As a result, the system effectively functions as a single composite system.

Under this configuration, the state and action spaces of the three arms are first unified. Specifically, the individual state spaces of the three arms are concatenated with the disk state to form a global state vector $s_{\text{all}} \in \mathbb{R}^{20}$, which represents the joint configurations of the three arms together with the disk height and tilt angle.

Similarly, the action spaces of the three arms are merged into a unified action space $a_{\text{all}} \in \mathbb{R}^{9}$, represented by a $9$-dimensional vector encoding the control signals of the nine actuated joints. This formulation is compatible with conventional single-agent reinforcement learning algorithms.

Under these settings, the SAC algorithm is employed for training. The actor network takes the 20-dimensional global state vector as input. This vector is processed by two fully connected layers, each consisting of 512 neurons with ReLU activation functions. The network then outputs the mean and standard deviation vectors of a Gaussian distribution, from which continuous actions are sampled. The sampled actions are subsequently passed through a $\texttt{tanh}$ activation function and an additional scaling operation to constrain the action range to $[-0.05, 0.05]$. The resulting 9-dimensional action vector is applied to the simulation environment as the control input.

The critic network takes as input the concatenation of the global state vector and the action vector. After being processed by two fully connected layers with ReLU activations, the network outputs a scalar value representing the estimated Q-value used for network updates. The overall input dimensionality of the critic is 29, and the output is a single scalar.

During training with the original SAC algorithm, a total of ten independent experiments were conducted. Under identical environmental settings, the parameters of both the actor and critic networks were randomly initialized at the beginning of each run, and each experiment was trained for $40,000$ episodes.

\begin{figure}
    \centering
    \includegraphics[width=0.45\textwidth]{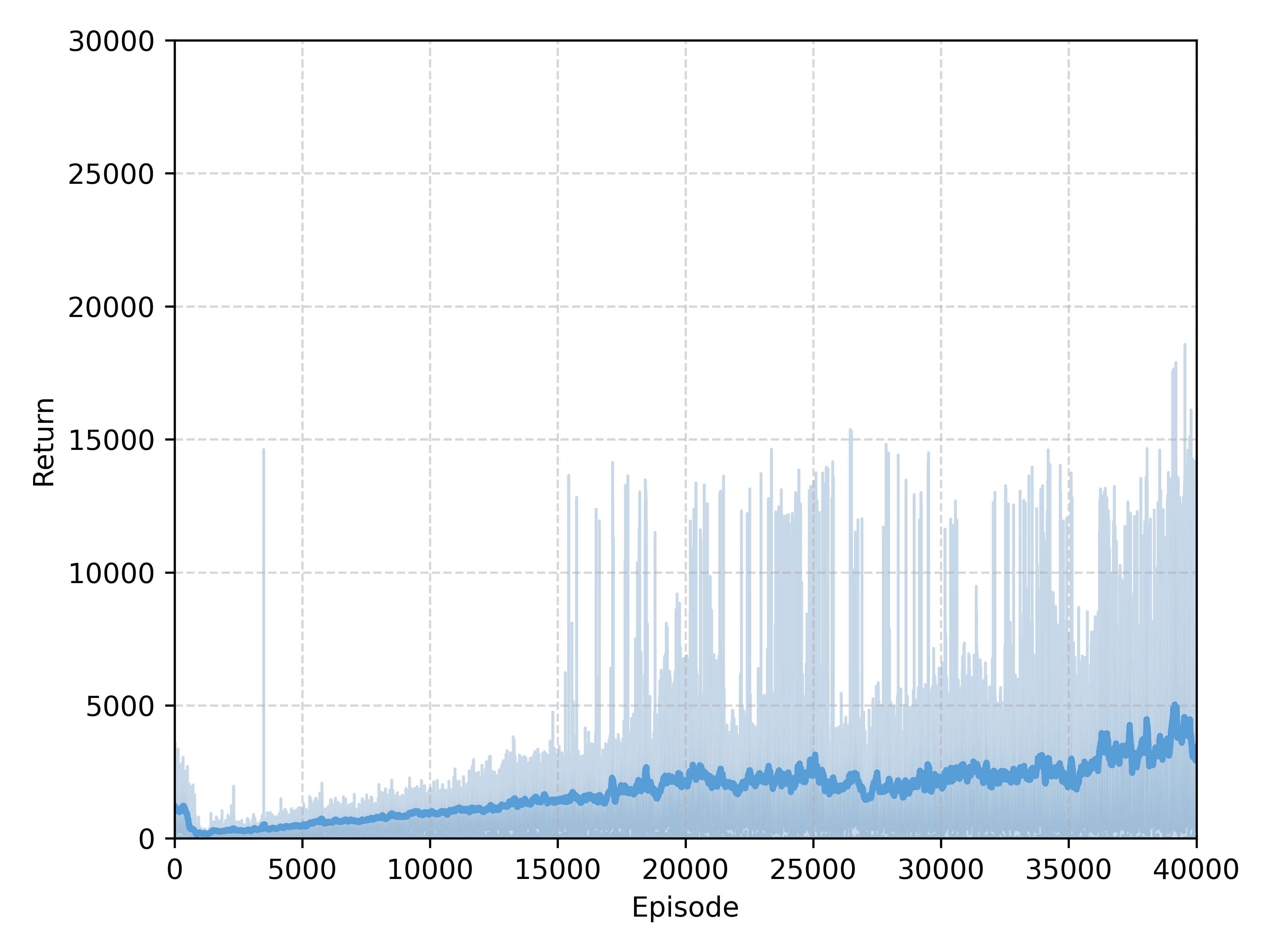}
    \caption{Performance of returns during SAC training.}
    \label{ONESimReturn}
 \end{figure}

The results show that, in all runs, the robotic arms successfully lifted the disk to a height of approximately $1.36\,\mathrm{m}$ while remaining within the defined safety constraints, as illustrated in Figure~\ref{ONESimReturn}. Hence, all experiments are considered successful.

\begin{figure}
    \centering
    \includegraphics[width=0.45\textwidth]{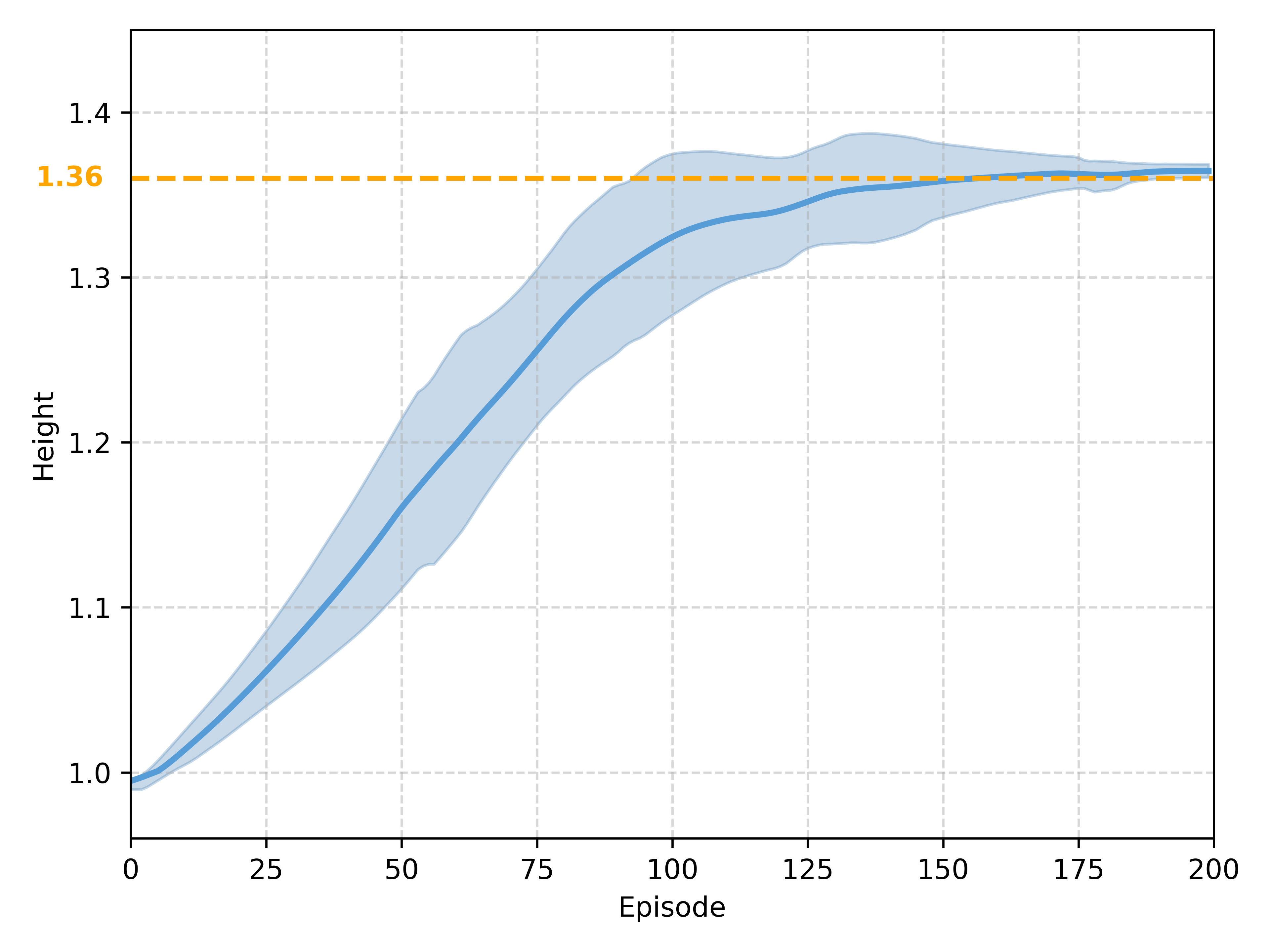}
    \caption{Performance of heights during SAC rendering (10 policies).}
    \label{ONESimHeightRender}
 \end{figure}
 
Finally, the trained policy was evaluated with both types of noise disabled. As shown in Figure~\ref{ONESimHeightRender}, under the control of the trained policy, the three-arm system successfully lifted the disk to a height close to the target level within a single task execution consisting of $200$ actions. These results demonstrate that the original SAC algorithm is capable of effectively accomplishing the specified collaborative lifting task.

\subsubsection{Decentralized Setting: CIEN-SAC}

In this experiment, no explicit communication is established among the robotic arms. Each arm updates its actions and optimizes its policy solely based on its information and observable variables, which are limited to the height and tilt angle of the disk. Under this decentralized setting, each arm is independently trained using the proposed CIEN-SAC algorithm.

For each CIEN-SAC instance, the actor network architecture follows the same design as in the original SAC configuration, with the only modification being an expanded input dimension of 10. Specifically, the input vector consists of the arm’s own 6-dimensional state, the 2-dimensional disk's state (height and tilt angle), and a 2-dimensional collective influence estimated by the CIEN. Since each arm is trained independently, the network scale is reduced compared to centralized SAC, with 256 neurons per hidden layer.

The CIEN adopts a variant of the DQN architecture, comprising two fully connected layers with 128 neurons each and ReLU activation functions, followed by a Tanh activation to constrain the output within the range [-1,1]. The input to this network is the 2-dimensional disk's state, and the output is a 2-dimensional estimate of the collective influence.

The critic network follows the same design as in the previous case; however, its input consists of the concatenation of multiple components, including the arm’s local state, the disk's state, the arm’s action, and the collective influence, resulting in a total input dimension of 13.

\begin{figure}
    \centering
    \includegraphics[width=0.45\textwidth]{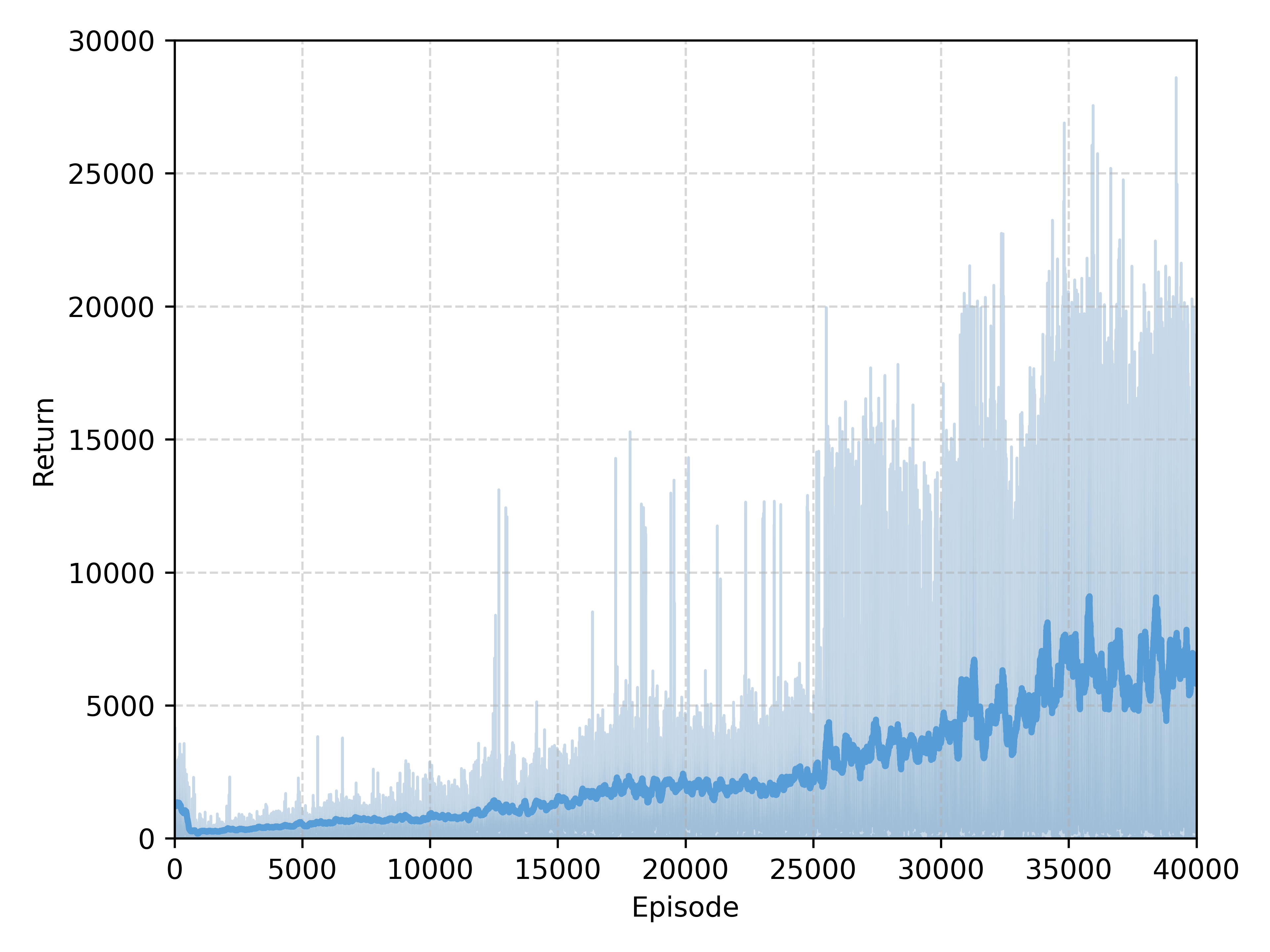}
    \caption{Performance of returns during CIEN-SAC training.}
    \label{ESTSimReturnGood}
\end{figure}

Similar to the ONE SAC setting, the CIEN-SAC algorithm is trained over ten independent runs. Due to the increased difficulty of coordinated exploration in the absence of inter-agent communication, the training process is slower than that of original SAC. As shown in Figure~\ref{ESTSimReturnGood}, nine out of ten runs successfully converge to effective policies within $40,000$ episodes. One run is recorded as a failure, where the learning process is too slow and the policy only achieves a lifting height of approximately $1.3\,\mathrm{m}$ by the end of training.

\begin{figure}
    \centering
    \includegraphics[width=0.45\textwidth]{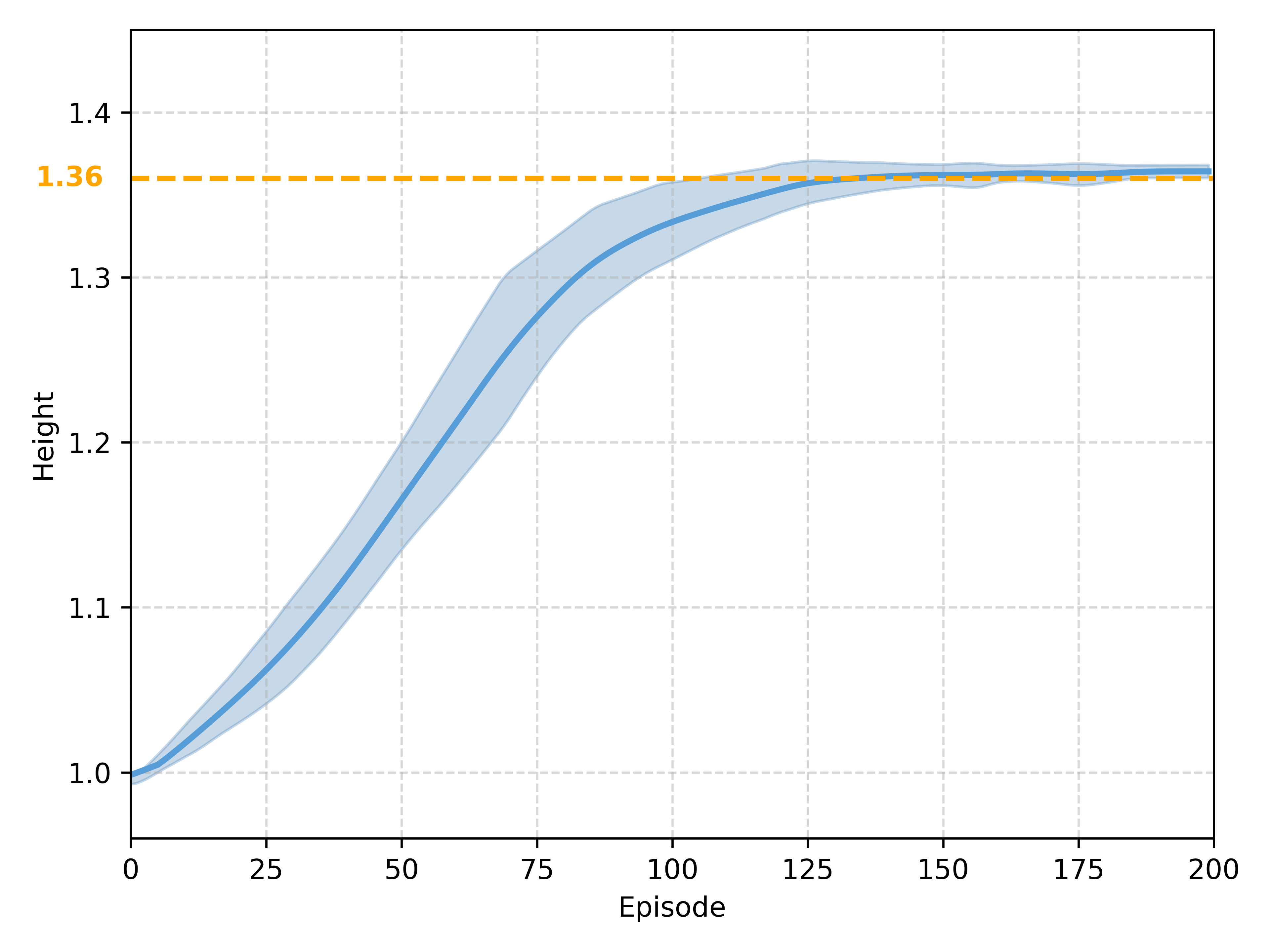}
    \caption{Performance of heights during CIEN-SAC rendering.}
    \label{ESTSimsHeightRender}
 \end{figure}

Figures~\ref{ESTSimsHeightRender} present the evaluation results of the learned policies after disabling exploration noise. The results demonstrate that the disk can be safely lifted to a height of approximately $1.36\,\mathrm{m}$, achieving performance comparable to that of centralized SAC. These comparative results validate the effectiveness of the proposed CIEN-SAC algorithm in decentralized settings, while retaining the advantages of communication-limited coordination.

\subsubsection{Decentralized Setting: SAC without No CIEN}

To confirm that the performance enhancement observed in CIEN-SAC is not solely attributable to the inclusion of the disk's state, we introduce decentralized SAC with out CIENas an ablation study. This experiment employs the same decentralized configuration as in CIEN-SAC, whereby no information sharing is permitted among the robotic arms. 

As the baseline for ablation, each agent is trained utilizing an independent SAC algorithm. The actor and critic network architectures for each agent are identical to those employed in the CIEN-SAC configuration, with the sole distinction being the complete removal of the CIEN. Consequently, policy learning depends solely on its local state and the observed state of the task object, without incorporating any form of collective influence modeling.

\begin{figure}
    \centering
    \includegraphics[width=0.45\textwidth]{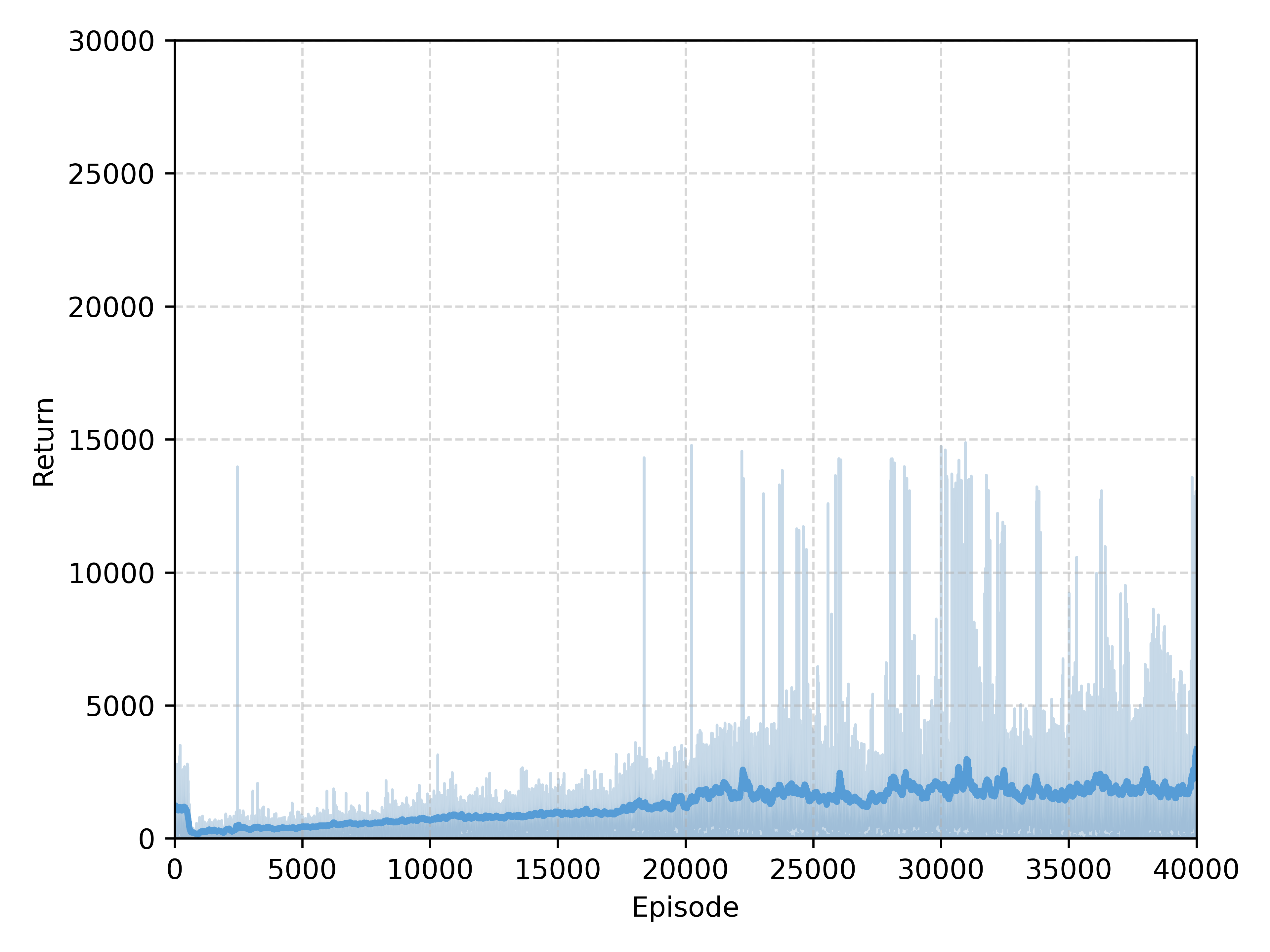}
    \caption{Performance of returns during independent SAC training.} 
    \label{NOESTSimsReturn}
\end{figure}

The independent SAC algorithm is also trained across ten separate runs. Due to the heightened complexity of coordinated exploration under communication limitations, coupled with the lack of collective influence modeling offered by CIEN, its overall training performance is significantly inferior to that of centralized SAC and CIEN-SAC. Following a total of $40,000$ training episodes, the learned policies demonstrate inadequate convergence, as indicated by the return curves presented in Figure~\ref{NOESTSimsReturn}.

These findings suggest that, within the three arm cooperative lifting task, independent SAC experiences markedly diminished convergence efficiency and task performance in environments with limited communication. In contrast, the superior performance of CIEN-SAC indicates that the observed enhancements in CIEN-SAC cannot be entirely attributed to the incorporation of disk's state information, thereby further confirming the essential role of CIEN in improving learning efficiency and stability in multiaagent cooperative environment.

\subsection{Policy Deployment and Experiments}

Although policies trained on simulation platforms such as Robosuite can attain stable and promising performance within simulation environment, the direct transfer of these policies to real robotic systems continues to pose significant challenges. Inevitable discrepancies between physical systems and simulation models, such as dynamics mismatches, actuator delays, control update frequencies, and mechanical constraints, frequently result in performance deterioration or even failure during real-world deployment, thereby establishing more rigorous demands on practical applicability and robustness.

To analyze and address these discrepancies, we further implement two representative policies demonstrating stable performance during simulation (centralized SAC and CIEN-SAC) on a physical robotic platform for validation. This section systematically details the procedure for sim-to-real policy transfer, outlines the construction and configuration of the physical experimental setup, and offers a comparative analysis of the experimental results obtained in simulation and real-world environments, with the objective of thoroughly evaluating the practical deployability of the proposed approach. 

The physical platform comprises three UR5e robotic arms, with hardware models, installation arrangements, and initial joint configurations precisely matching those employed in the simulation environment, thereby reducing the impact of extraneous structural variations on the experimental results.

\subsubsection{Vision-Based Task Object's State Estimation}

Within the RoboSuite environment, the state of each object can be directly obtained from the environment. However, this is not feasible in the real-world environment. In physical environments, each arm can only obtain its own joint state and cannot acquire the state of the disk. Therefore, additional methods are required to obtain this value.

Given that the proposed algorithm's advantage lies in communication requirements, this study employs a vision-based benchmark marker tracking method to preserve this benefit while acquiring the disk's state. 

In this method, QR code-style benchmark markers affixed to the disk's surface enable the camera to reliably detect and localize the markers under diverse lighting and viewing conditions. During operation, a calibrated RGB camera continuously observes the marker points and extracts their corner features. By solving the perspective-in-perspective (PnP) problem from the detected marker geometry, the object's full 6-degree-of-freedom pose—including its 3D position and orientation—is reconstructed in real time. Consequently, any detected pose change directly reflects variations in the disk's height and rotational state. This approach provides a stable source of disk's state information, which can be further transformed into a disk's state representation consistent with that used in the simulation environment. The resulting state estimates can then be seamlessly integrated into the networks or control policies for collaborative manipulation, thereby supporting effective deployment in real-world systems.

\subsubsection{Training with Observation Noise}

As mentioned earlier, we use a camera to observe QR codes to estimate the disk's state. However, during actual deployment, we discovered that the information state of the disk obtained through this method contains errors. Specifically, the height and angle measurements exhibit errors of $1 cm$ and $1 degree (angular value)$, respectively. When fed as state inputs to the actor neural network, these errors are amplified, leading to significant inaccuracies in the generated actions. This error resulted in poor performance during the initial phase of transferring from the simulation environment to the real environment, causing the gripper to repeatedly squeeze the disk.

Since such noise is difficult to completely eliminate, we further introduce a complementary strategy built upon the existing approach. This ensures the trained policy possesses sufficient robustness to overcome the impact of these errors. Based on this approach, we selected one successful policy each from the strategies trained using the centralized SAC and CIEN-SAC algorithms in the Robosuite environment. Each selected policy trained an additional 8,000 episodes. Since observation noise significantly impacts training efficiency, we appropriately reduced the task difficulty in this phase, having already validated the effectiveness of our proposed algorithm in the simulation environment. Specifically, while maintaining the safety conditions, the task target height was reduced from $1.36m$ to $1.25m$.

By inspecting the learning curves, it can be observed that CIEN-SAC achieves a substantially higher average return after approximately $4,000$ episodes, indicating superior training performance under random initialization and stochastic perturbation noise. This improvement can be attributed to the fact that, in contrast to centralized SAC, where the three robotic arms are treated as a single agent governed by a unified policy network, CIEN-SAC introduces independent stochastic noise into the observations of each arm, thereby significantly enhancing policy diversity and robustness.

The presence of independent noise exposes each robotic arm to a richer and more heterogeneous state distribution during training, preventing the learned policy from over-relying on idealized, synchronized observation patterns. Such broader state-space coverage not only improves the policy’s generalization capability under various environmental disturbances, but also mitigates several issues commonly associated with centralized training, including overly tight policy coupling and heightened sensitivity to observation bias. As a result, the learned policy exhibits improved stability and adaptability in collaborative settings.

\subsubsection{Results and Evaluation}

\begin{figure}
    \centering
    \includegraphics[width=0.45\textwidth]{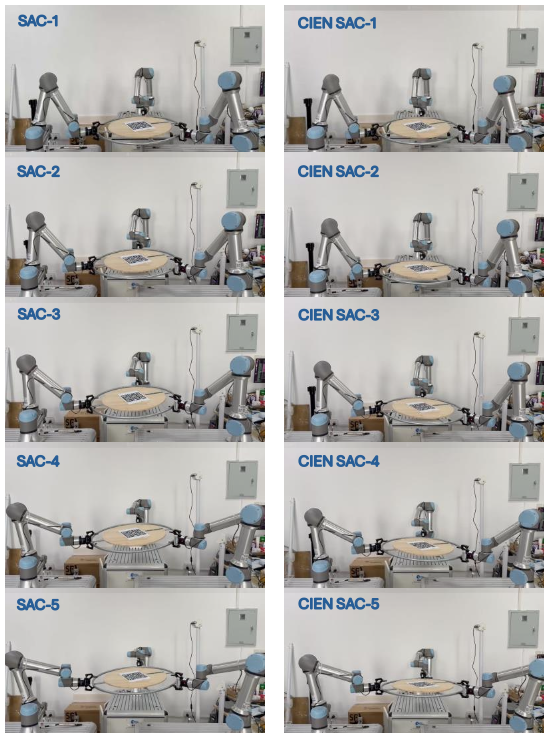}
    \caption{Sequential snapshots of using TD3 and AENTD3 in physical robotic platform.}
    \label{Snapshots}
\end{figure}

\begin{figure}
    \centering
    \includegraphics[width=0.45\textwidth]{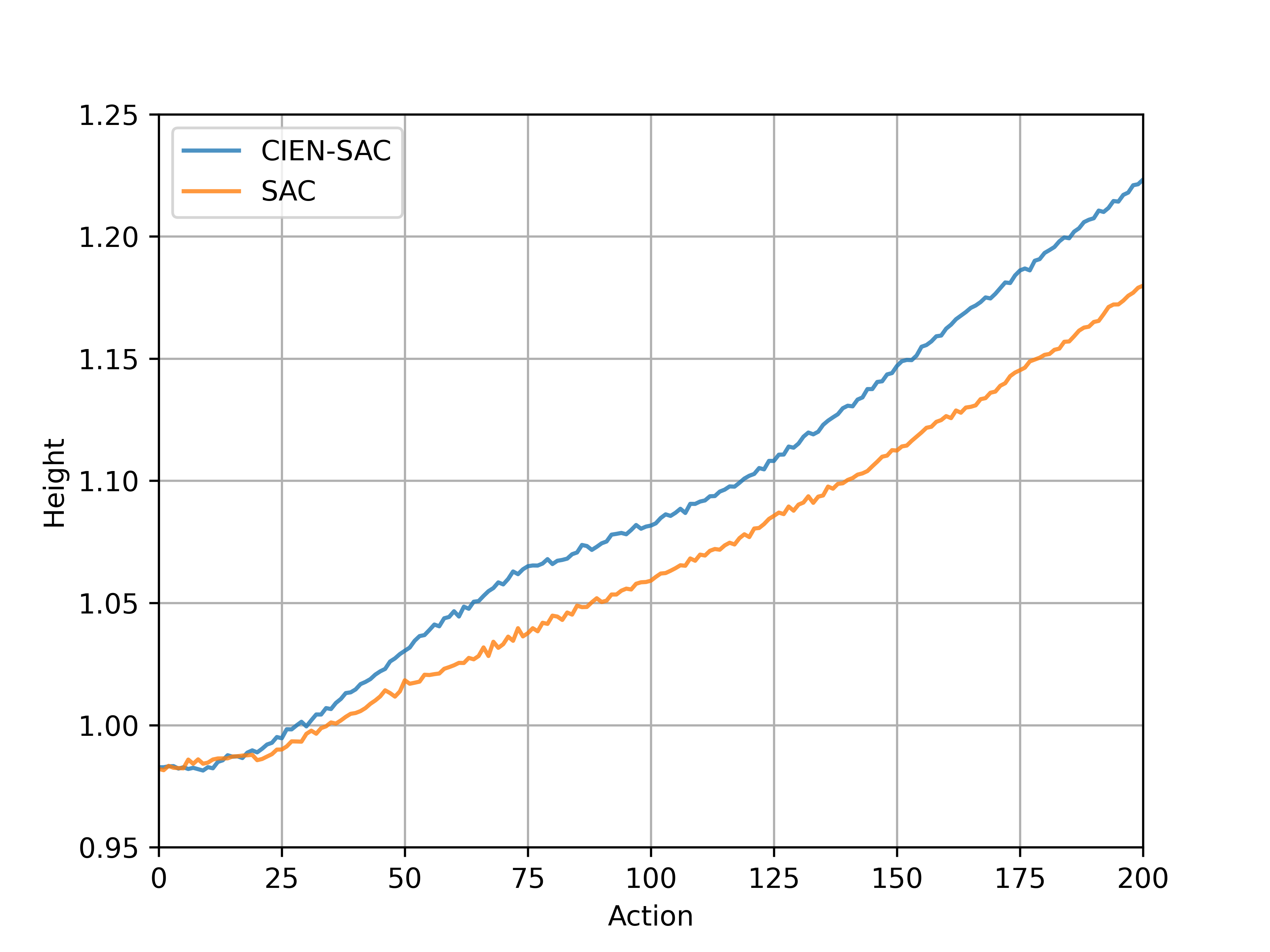}
     \caption{Height trajectories of the disk in physical robotic platform.}
    \label{RH}
\medskip
    \centering
     \includegraphics[width=0.45\textwidth]{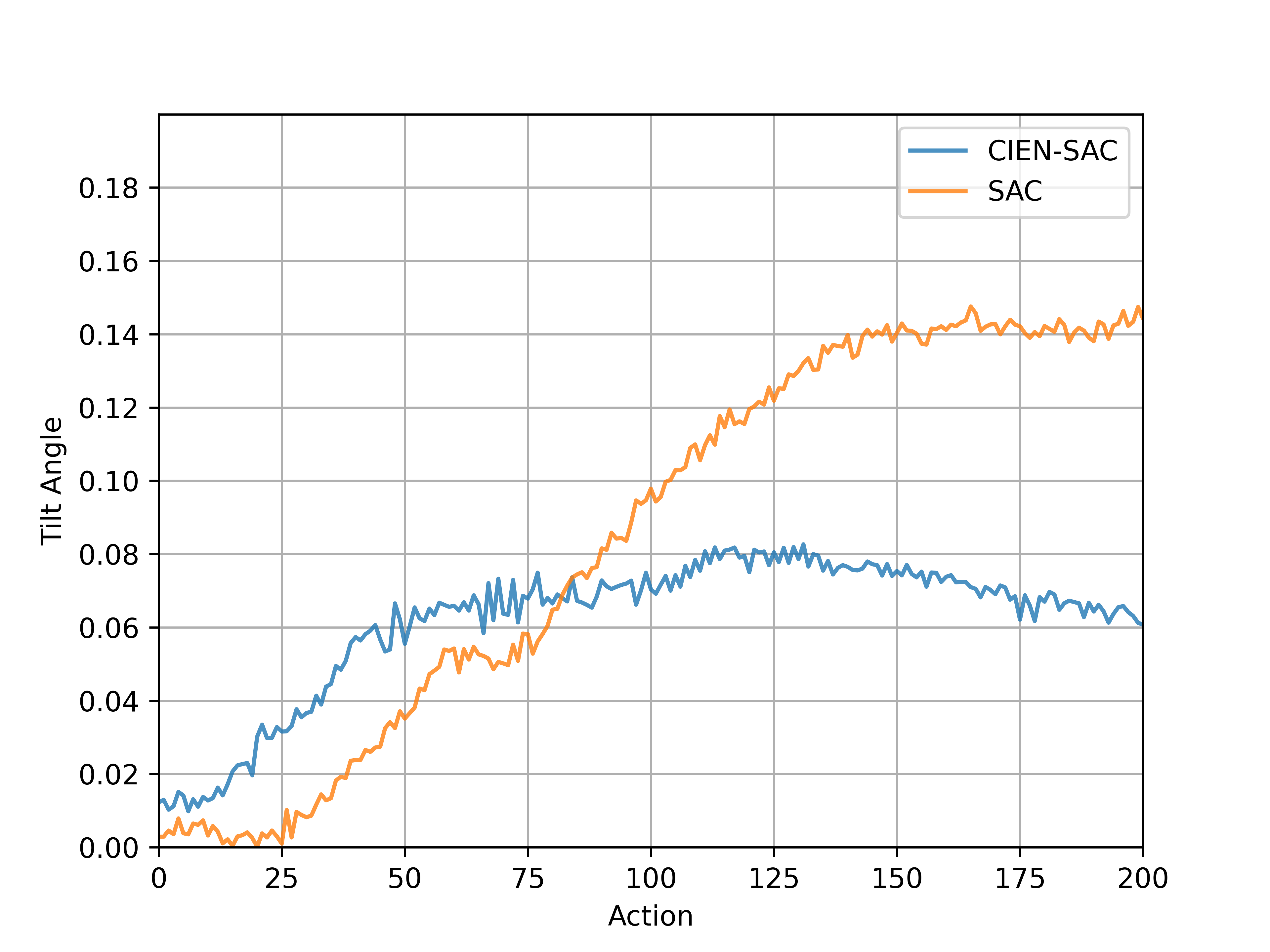}
    \caption{Tilt angle trajectories of the disk in physical robotic platform.}
    \label{RA}
\end{figure}

After deploying the converged centralized SAC and CIEN-SAC policies to the physical robotic platform, their performance was evaluated through real-world lifting experiments. The entire lifting process lasted approximately 12 seconds. To visually illustrate the task execution, representative snapshots were captured at 3 second intervals, as shown in Figure~\ref{Snapshots}. The left and right columns correspond to the execution results of the centralized SAC and CIEN-SAC policies, respectively. Experimental observations indicate that both methods successfully accomplished the disk lifting task while maintaining stable operation throughout the process, with no noticeable squeezing or structural damage observed. Furthermore, Figures~\ref{RH} and~\ref{RA} present the height and tilt angle trajectories achieved in the real environment, providing quantitative evidence for performance comparison between the two approaches.

Experimental results further demonstrate that, under communication-limited conditions, where the three robotic arms operate as independent agents without access to each other’s action information, the proposed CIEN is still capable of inferring task relevant latent information from local observations. Consequently, CIEN-SAC achieves task performance comparable to that of the centralized SAC algorithm, despite the absence of full information sharing during both training and execution. In addition, experiments incorporating observation noise reveal that the CIEN framework exhibits strong robustness and disturbance tolerance. These findings indicate that the proposed method is well suited for multiagent cooprative manipulation under communication-limited and shows strong potential for real-world deployment.

\section{Conclusion}
\label{sec.Conclusion}

This paper addresses scalable multiagent reinforcement learning for cooperative manipulation under communication-limited and partial observability by proposing a SAC-based framework augmented with a Collective Influence Estimation Network (CIEN). By explicitly modeling the collective influence among agents and inferring critical interaction information solely from local observations, the proposed approach enables three robotic arms to achieve stable and coordinated lifting behavior without relying on inter-agent communication, making it well suited for scalable multi-agent settings. Experimental results demonstrate that, even in the absence of complete information sharing, CIEN-SAC achieves task performance comparable to centralized SAC baselines, while exhibiting enhanced stability in the presence of observation noise.

Furthermore, policies trained under the collective influence modeling framework are deployed on a physical robotic platform and systematically compared with policies learned under centralized assumptions. Real-world experimental results indicate that the proposed method maintains reliable and feasible execution performance, highlighting its practical scalability and applicability in complex collaborative manipulation scenarios. Future work will focus on improving training efficiency for large-scale multiagent systems, extending the collective influence estimation framework to non-cooperative and mixed-interaction tasks, and further investigating its scalability to scenarios involving a greater number of agents.

\bibliographystyle{IEEEtran}
\bibliography{reference}

@article{luo2025multi,
  title={Multi-Agent Reinforcement Learning with Deep Networks for Diverse Q Q-Vectors},
  author={Luo, Zhenglong and Chen, Zhiyong and Liu, Shijian and Welsh, James},
  journal={Electronics Letters},
  volume={61},
  number={1},
  pages={e70342},
  year={2025},
  publisher={Wiley Online Library}
}

@article{watkins1992q,
  title={Q-Learning},
  author={Watkins, Christopher JCH and Dayan, Peter},
  journal={Machine Learning},
  volume={8},
  pages={279--292},
  year={1992},
  publisher={Springer}
}

@article{lecun2015deep,
  title={Deep Learning},
  author={LeCun, Yann and Bengio, Yoshua and Hinton, Geoffrey},
  journal={Nature},
  volume={521},
  number={7553},
  pages={436--444},
  year={2015},
  publisher={Nature Publishing Group UK London}
}

@article{hu2003nash,
  title={Nash Q-learning for General-Sum Stochastic Games},
  author={Hu, Junling and Wellman, Michael P},
  journal={Journal of Machine Learning Research},
  volume={4},
  number={Nov},
  pages={1039--1069},
  year={2003}
}

@inproceedings{todorov2012mujoco,
    title={Mujoco: A Physics Engine for Model-Based Control},
    author={Todorov, Emanuel and Erez, Tom and Tassa, Yuval},
    booktitle={2012 IEEE/RSJ International Conference on Intelligent Robots and Systems},
    pages={5026--5033},
    year={2012},
    organization={IEEE}
    }

@article{lowe2017multi,
  title={Multi-agent Actor-Critic for Mixed Cooperative-Competitive Environments},
  author={Lowe, Ryan and Wu, Yi I and Tamar, Aviv and Harb, Jean and Pieter Abbeel, OpenAI and Mordatch, Igor},
  journal={Advances in Neural Information Processing Systems},
  volume={30},
  year={2017}
}

@article{hausknecht2015deep,
  title={Deep Recurrent Q-Learning for Partially Observable MDPs},
  author={Hausknecht, Matthew and Stone, Peter},
  journal={arXiv preprint arXiv:1507.06527},
  year={2015}
}

@inproceedings{foerster2018counterfactual,
  title={Counterfactual Multi-Agent Policy Gradients},
  author={Foerster, Jakob and Farquhar, Gregory and Afouras, Triantafyllos and Nardelli, Nantas and Whiteson, Shimon},
  booktitle={Proceedings of the AAAI Conference on Artificial Intelligence},
  volume={32},
  number={1},
  year={2018}
}

@article{zhu2020robosuite,
title={Robosuite: A Modular Simulation Framework and Benchmark for Robot Learning},
author={Zhu, Yuke and Wong, Josiah and Mandlekar, Ajay and Mart{\'\i}n-Mart{\'\i}n, Roberto},
journal={arXiv preprint arXiv:2009.12293},
year={2020}
}

@inproceedings{fujimoto2018td3,
  title={Addressing Function Approximation Error in Actor-Critic Methods},
  author={Fujimoto, Scott and Hoof, Herke and Meger, David},
  booktitle={International Conference on Machine Learning (ICML)},
  year={2018}
}

@article{tan1993iql,
  title={Multi-Agent Reinforcement Learning: Independent vs. Cooperative Agents},
  author={Tan, Ming},
  journal={Proceedings of the Tenth International Conference on Machine Learning (ICML)},
  year={1993}
}

@inproceedings{foerster2018lola,
  title={Learning with Opponent-Learning Awareness},
  author={Foerster, Jakob and Chen, Richard and Al-Shedivat, Maruan and Whiteson, Shimon and Abbeel, Pieter and Mordatch, Igor},
  booktitle={Proceedings of the 17th International Conference on Autonomous Agents and MultiAgent Systems (AAMAS)},
  year={2018},
  pages={122–130}
}

@inproceedings{foerster2016dial,
  title={Learning to communicate with deep multi-agent reinforcement learning},
  author={Foerster, Jakob N and Assael, Yannis M and de Freitas, Nando and Whiteson, Shimon},
  booktitle={Advances in Neural Information Processing Systems},
  volume={29},
  pages={2137--2145},
  year={2016}
}

@inproceedings{sukhbaatar2016commnet,
  title={Learning multiagent communication with backpropagation},
  author={Sukhbaatar, Sainbayar and Fergus, Rob and others},
  booktitle={Advances in Neural Information Processing Systems},
  volume={29},
  pages={2244--2252},
  year={2016}
}

@inproceedings{jiang2020graph,
  title={Graph Convolutional Reinforcement Learning},
  author={Jiang, Jiechuan and Dun, Baoxiang and Yang, Tianyu and Lu, Zongzhang},
  booktitle={International Conference on Learning Representations (ICLR)},
  year={2020}
}

@inproceedings{haarnoja2018sac,
  title={Soft Actor-Critic: Off-Policy Maximum Entropy Deep Reinforcement Learning with a Stochastic Actor},
  author={Haarnoja, Tuomas and Zhou, Aurick and Abbeel, Pieter and Levine, Sergey},
  booktitle={International Conference on Machine Learning (ICML)},
  year={2018}
}

@inproceedings{yang2020rommeo,
  title={A Regularized Framework for Opponent Modeling in Multi-Agent Reinforcement Learning},
  author={Yang, Yiding and Wang, Jun and Xu, Yixuan and Wang, Yaodong},
  booktitle={International Conference on Machine Learning (ICML)},
  year={2020}
}

@inproceedings{albrecht2018bolo,
  title={Bayesian Opponent Learning in Multi-Agent Systems},
  author={Albrecht, Stefano V and Stone, Peter},
  booktitle={Advances in Neural Information Processing Systems},
  volume={31},
  year={2018}
}

@inproceedings{du2021omtcn,
  title={Opponent Modeling with Temporal Convolutional Networks in Multi-Agent Reinforcement Learning},
  author={Du, Yali and Ding, Zilong and Zhang, Hao and Wang, Jun},
  booktitle={International Joint Conference on Artificial Intelligence (IJCAI)},
  year={2021}
}

@inproceedings{hong2018dpiqn,
  title={Deep Policy Inference Q-Network for Multi-Agent Systems},
  author={Hong, Zhang-Wei and Shann, Tzu-Yun and Su, Shih-Yang and Chang, Yi-Hsiang and Lin, Chung-Yi},
  booktitle={International Conference on Learning Representations (ICLR)},
  year={2018}
}

@inproceedings{hong2019drpiqn,
  title={Recursive Reasoning for Multi-Agent Reinforcement Learning},
  author={Hong, Zhang-Wei and Su, Shih-Yang and Chang, Yi-Hsiang and Shann, Tzu-Yun},
  booktitle={International Conference on Learning Representations (ICLR)},
  year={2019}
}

@inproceedings{wen2019pr2,
  title={Probabilistic Recursive Reasoning for Multi-Agent Reinforcement Learning},
  author={Wen, Ying and Yang, Yaodong and Wang, Jun and Pan, Wei},
  booktitle={International Conference on Learning Representations (ICLR)},
  year={2019}
}

@inproceedings{zhang2021maop,
  title={Multi-Agent Offline Policy Inference},
  author={Zhang, Ruohan and Geng, Zhaoyang and Xu, Huazhe and Wang, Jun},
  booktitle={Advances in Neural Information Processing Systems},
  volume={34},
  year={2021}
}

@inproceedings{wang2020i2q,
  title={Imitated Opponent Q-Learning for Multi-Agent Reinforcement Learning},
  author={Wang, Yaodong and Hao, Jianye and Wang, Jun},
  booktitle={International Conference on Learning Representations (ICLR)},
  year={2020}
}

@inproceedings{yang2018meanfield,
  title={Mean Field Multi-Agent Reinforcement Learning},
  author={Yang, Yaodong and Luo, Ruihan and Li, Ming and Zhou, Ming and Zhang, Weinan and Wang, Jun},
  booktitle={Proceedings of the 35th International Conference on Machine Learning (ICML)},
  year={2018}
}

@article{luo2026aen,
  title={Multiagent Reinforcement Learning with Neighbor Action Estimation},
  author={Luo, Zhenglong and Chen, Zhiyong and Liu, Aoxiang},
  journal={arXiv preprint arXiv:2601.04511},
  year={2026},
  doi={10.48550/arXiv.2601.04511},
  primaryClass={cs.RO}
}

@article{KondaTsitsiklis2000,
  title   = {Actor-Critic Algorithms},
  author  = {Konda, Vijay R. and Tsitsiklis, John N.},
  journal = {Advances in Neural Information Processing Systems},
  volume  = {12},
  year    = {2000}
}

\end{document}